\documentclass[letterpaper]{article}           
\usepackage{aaai2026}              
\usepackage[utf8]{inputenc}
\usepackage[T1]{fontenc}

\usepackage[most]{tcolorbox} 
\usepackage{xcolor}          

\usepackage{times}     
\usepackage{helvet}    
\usepackage{courier}   
\usepackage[hyphens]{url}      
\usepackage{graphicx}           
\usepackage{float}  
\usepackage{amsmath}
\usepackage{algorithm}
\usepackage{algpseudocode}
\usepackage{textcomp}       

\usepackage{amssymb}
\usepackage{natbib}
\usepackage{caption}            
\usepackage{algorithm}      
\usepackage{algpseudocode}  
\usepackage{booktabs}
\usepackage{tabularx}

\usepackage{booktabs}       
\usepackage[table]{xcolor}  
\usepackage{colortbl}
\usepackage[inline]{enumitem}  
\usepackage{pifont}         
\usepackage{newfloat}
\usepackage{listings}
\DeclareCaptionStyle{ruled}{labelfont=normalfont,labelsep=colon,strut=off}
\floatstyle{ruled}
\newfloat{listing}{tb}{lst}{}
\floatname{listing}{Listing}

\title{Top-Down Semantic Refinement for Image Captioning}

\author{
Jusheng Zhang\textsuperscript{\rm 1},
Kaitong Cai\textsuperscript{\rm 1},
Jing Yang\textsuperscript{\rm 1},
Jian Wang\textsuperscript{\rm 1},
Chengpei Tang\textsuperscript{\rm 1},
Keze Wang\textsuperscript{\rm 1}\thanks{Corresponding Author}
}

\affiliations{
\textsuperscript{\rm 1}Sun Yat-sen University \\
}

\begin{document}

\maketitle
\begin{abstract}
Large Vision-Language Models (VLMs) face an inherent contradiction in image captioning: their powerful single-step generation capabilities often lead to a \textbf{myopic} decision-making process. This makes it difficult to maintain global narrative coherence while capturing rich details, a limitation that is particularly pronounced in tasks that require multi-step and complex scene description. To overcome this fundamental challenge, we redefine image captioning as a \textbf{goal-oriented hierarchical refinement planning problem}, and further propose a novel framework, named Top-Down Semantic Refinement (TDSR), which models the generation process as a Markov Decision Process (MDP). However, planning within the vast state space of a VLM presents a significant computational hurdle. Our core contribution, therefore, is the design of a \textbf{highly efficient Monte Carlo Tree Search (MCTS) algorithm tailored for VLMs}. By incorporating a \textbf{visual-guided parallel expansion} and a \textbf{lightweight value network}, our TDSR reduces the call frequency to the expensive VLM by an order of magnitude without sacrificing planning quality. Furthermore, an adaptive early stopping mechanism dynamically matches computational overhead to the image's complexity. Extensive experiments on multiple benchmarks, including DetailCaps, COMPOSITIONCAP, and POPE, demonstrate that our TDSR, as a plug-and-play module, can significantly enhance the performance of existing VLMs (e.g., LLaVA-1.5, Qwen2.5-VL) by achieving state-of-the-art or highly competitive results in fine-grained description, compositional generalization, and hallucination suppression.
\end{abstract}

\section{Introduction}
\label{sec:intro}
At the intersection of computer vision and natural language processing, Large Vision-Language Models (VLMs)\cite{clip,clip2,clip3,clip4,clip4zs,Attention,M10,M11,MAB} have become the dominant force in image captioning. Through powerful visual encoders and language decoders, these models can generate fluent text that is generally aligned with the image content\cite{BLIP,BLIP-2,git2,clip,Z3}. However, despite their remarkable success, the core auto-regressive generation mechanism of VLMs exposes a fundamental flaw, i.e., an inherent lack of planning capability. When generating each token, VLMs typically employ greedy or beam search strategies\cite{brown2020languagemodelsfewshotlearners,ICML,ICML2,radford2019language,8099614,Z8}. This decision-making process is inherently ``myopic", confined to maximizing local probabilities without ``deliberate thought" or foresight and planning capability for the global narrative structure.

This lack of planning capability leads to an intractable dilemma: the model either produces a coherent but detail-poor ``safe" description to ensure consistency, or it generates factual errors and logical breaks, i.e., the ``hallucination" phenomenon, when attempting to capture rich details without global guidance \cite{clip2,clip,rohrbach2019objecthallucinationimagecaptioning,openai2023gpt4v,Z7,z20}. To address this challenge, the research community once turned to a seemingly intuitive ``bottom-up" paradigm\cite{zdes,zdes2,zdes3,zdes4,zdes5}. These methods first detect independent regions in an image, describe them separately, and finally ``stitch" these fragmented descriptions into a complete caption. However, this ``local-to-global" strategy fails to address the core problem. Lacking a unified global plan as an anchor from the outset, the resulting descriptions often degenerate into a simple list of facts, leading to semantic fragmentation and logical incoherence\cite{rohrbach2019objecthallucinationimagecaptioning,lianguan,bugliarello2021rolesyntacticplanningcompositional,zdes5zs,Z2}. This proves that merely stitching details together cannot effectively compensate for the VLM's lack of planning ability.

\begin{figure}[!t]
    \centering
    \hspace*{-0.40cm}
    \includegraphics[scale=0.22]{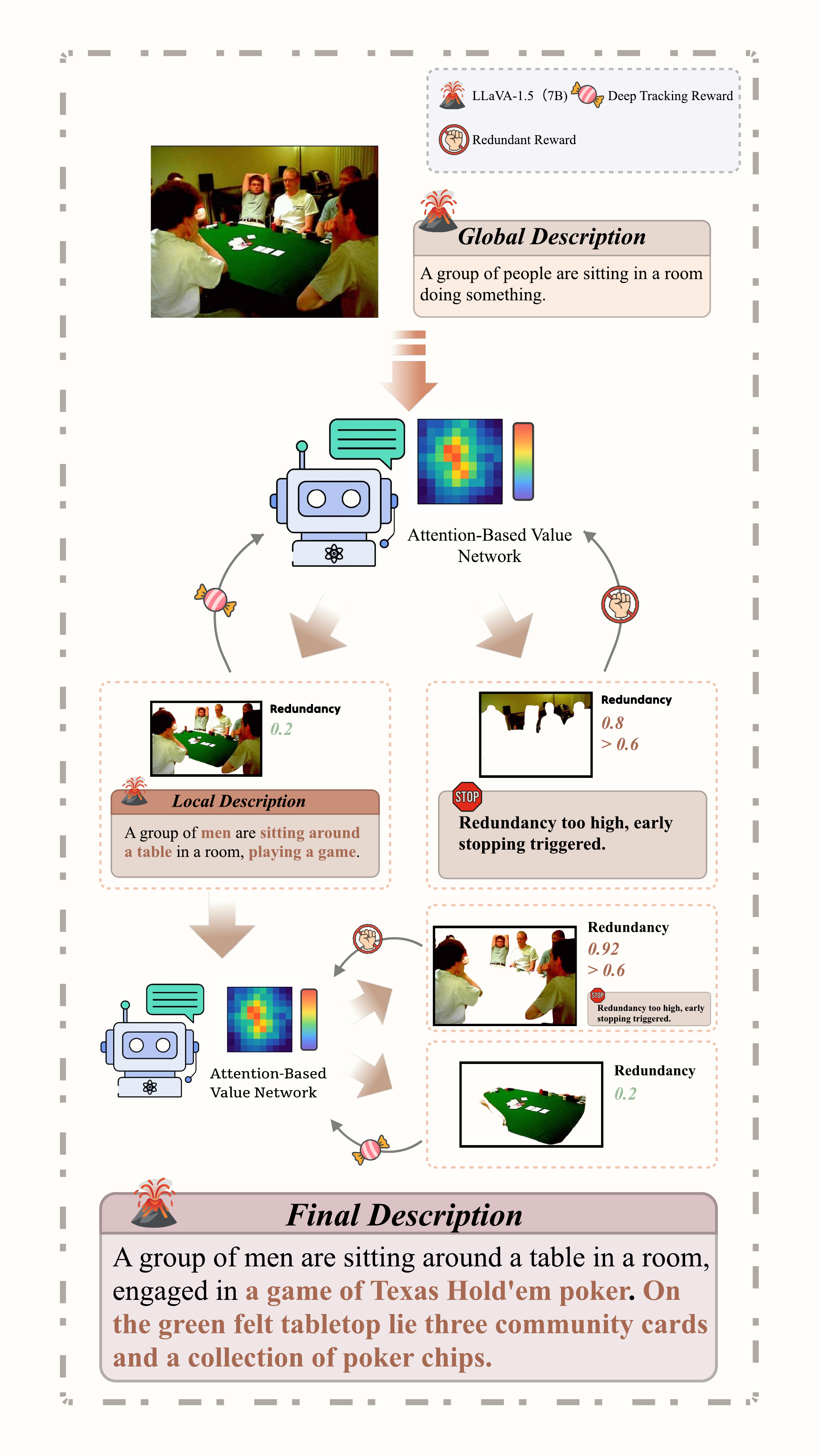}
    \caption{The TDSR framework generates coherent, detailed captions through global-to-local refinement, guided by redundancy-aware stopping and efficient MCTS.}
    \label{fig:shouyetu111}
\end{figure}

We argue that the root of the problem lies in the generation paradigm itself, and the solution is to fundamentally reframe image captioning as a planning problem. To this end, we propose an innovative ``top-down" semantic refinement framework (TDSR), which redefines the task from a unidirectional generation process into a coarse-to-fine, goal-oriented, hierarchical planning process\cite{yao2019planandwritebetterautomaticstorytelling,yarats2018hierarchicaltextgenerationplanning,Z1}. This idea, illustrated in Figure~\ref{fig:shouyetu111}, mimics the human cognitive process\cite{marr1982vision,bar2013topdown,mefford2023varied}: first, form a holistic impression of the image to generate a high-level, core description as a ``planning blueprint" (e.g., for a picture of people playing cards, an initial description might be ``a group of people are sitting in a room doing something"). Then, using this blueprint as a guide, purposefully and progressively explore and fill in key details (e.g., further specifying it as ``a group of men are sitting around a table, engaged in a game of Texas Hold'em poker," and adding that ``on the green felt tabletop lie three community cards and a collection of poker chips"). This ``global guidance, local refinement" mechanism ensures that all details serve a unified narrative goal, fundamentally guaranteeing high coherence and richness in the description.

Translating this elegant planning concept into an effective computational process hinges on efficient search and planning within the vast language space\cite{hoang2017decodingcontinuousoptimizationneural,wiher2022decodingstrategiesneuraltext}. We rigorously formalize this process as a Markov Decision Process (MDP\cite{puterman1994mdp}) and employ Monte Carlo Tree Search (MCTS\cite{_wiechowski_2022,Kemmerling_2023}) as the core engine. However, directly applying standard MCTS to a VLM is computationally infeasible due to the model's massive inference cost\cite{6145622}. Therefore, our core technical contribution lies in deeply optimizing the MCTS algorithm to enable efficient planning within VLMs. By incorporating a Visual-Guided Parallel Expansion mechanism and a lightweight value network, our algorithm reduces the call frequency to the expensive VLM by an order of magnitude without sacrificing planning quality, successfully resolving the efficiency bottleneck. Our framework (TDSR), as a plug-and-play module, significantly enhances the performance of existing VLMs and achieves state-of-the-art or highly competitive results on multiple benchmarks.

Our core contributions can be summarized in three points:
\begin{itemize}
    \item \textbf{A Novel ``Planning-based" Generation Paradigm:} We propose a ``Top-Down" planning framework (TDSR) that redefines image captioning as a coarse-to-fine hierarchical planning problem. This fundamentally resolves the ``myopic" flaw of traditional generative models, ensuring both global narrative coherence and local detail richness.

    \item \textbf{An Efficient MCTS Algorithm Tailored for VLMs:} We design a highly efficient Monte Carlo Tree Search (MCTS) algorithm to address the high inference cost of VLMs. The algorithm broadens search breadth via a ``Visual-Guided Parallel Expansion" mechanism and uses a ``lightweight value network" for fast value estimation, reducing the call frequency to the expensive VLM by an order of magnitude without sacrificing planning quality.

    \item \textbf{A Dynamic and Adaptive Search Control Strategy:} We introduce a dynamic control strategy to enhance planning efficiency and quality. The strategy guides the search direction through a composite reward function combining a ``redundancy penalty" and a ``depth incentive," and it intelligently allocates computational resources based on image complexity via an ``adaptive early stopping" mechanism to avoid unnecessary overhead.
\end{itemize}

\section{Related Work}
\label{sec:related_work}
\subsection{Early Encoder-Decoder Architectures}
As a cross-disciplinary field between computer vision and natural language processing\cite{zhang2025cfvlmcounterfactualvisionlanguagefinetuning,clip,clip2,clip4zs,Z4,SGN}, early research in image captioning primarily adopted the encoder-decoder framework. Methods represented by Show and Tell~\cite{7298935,MMCOT,HTC} used a CNN\cite{krizhevsky2012imagenet,726791,M1,M2,M3,M4} to extract global image features and an RNN\cite{elman1990finding,M5,M6,M7,M9,M8} to generate a fluent description. Subsequently, Show, Attend and Tell~\cite{7298935} introduced an attention mechanism, allowing the model to focus on local regions of the image. These pioneering works excelled at generating grammatically coherent sentences, but their mechanism of generating a single, holistic description meant they often overlooked fine-grained object details, leading to a lack of richness and specificity. \textbf{This is precisely one of the core problems our TDSR framework aims to solve through multi-step refinement.}
\subsection{The ``Bottom-up" Generation Paradigm}
To enhance detail capture, subsequent research shifted towards the ``bottom-up" paradigm\cite{zdes,zdes2,zdes3,zdes4,zdes5}. This approach typically first identifies independent objects or regions in an image, generates local descriptions for them, and finally stitches these segments into a complete sentence. DenseCap~\cite{DenseCap} is a typical representative of this line of work, which utilizes an object detector to locate and describe regions individually. Follow-up works, such as Patch Matters~\cite{Patch,RAN,POT} and FineCaption~\cite{FINECAPTION}, focused on improving the quality of these local descriptions. Although these methods significantly increased detail richness, their ``split-first, stitch-later" process inherently decouples from the global context, often leading to semantic fragmentation and insufficient global coherence. In stark contrast, our ``top-down" approach, guided by global context, fundamentally avoids the inconsistency issues inherent in the ``split-first, stitch-later" process.

\subsection{Large Vision-Language Models and Generation Refinement Strategies}
In recent years, Large Vision-Language Models (VLMs), such as LLaVA-1.5~\cite{llava}, Qwen-VL~\cite{Qwen}, and Ferret~\cite{you2023ferretrefergroundgranularity}, have significantly advanced visual narrative capabilities through pre-training on massive image-text data. However, despite their powerful foundational abilities, their standard autoregressive generation still faces the inherent trade-off between detail and coherence. Consequently, several training-free enhancement methods have emerged. For instance, some works employ iterative prompting (e.g., IT~\cite{zhou2023leasttomost}) to induce the model to output more details.

A more promising direction involves formalizing the generation process as a search problem and employing planning algorithms like Monte Carlo Tree Search (MCTS) for optimization~\cite{MCTS,yao2023treethoughtsdeliberateproblem,silver2017masteringchessshogiselfplay}. Against this backdrop, our TDSR framework proposes a more fundamental solution. It also employs MCTS, but its core innovation lies in how to tailor and efficiently execute the search for VLMs. Our method is guided by a sophisticated composite reward function and integrates a suite of efficiency optimization mechanisms, including visual-guided parallel rollouts and dynamic redundancy control. \textbf{Consequently, TDSR is not only superior in its paradigm ('top-down') but also innovative in its solution strategy (efficient MCTS), thereby efficiently unifying detail and coherence while significantly mitigating issues like semantic fragmentation and content hallucination.}
\begin{figure*}[h]
    \centering
    \hspace*{-0.40cm}
    \includegraphics[scale=0.24]{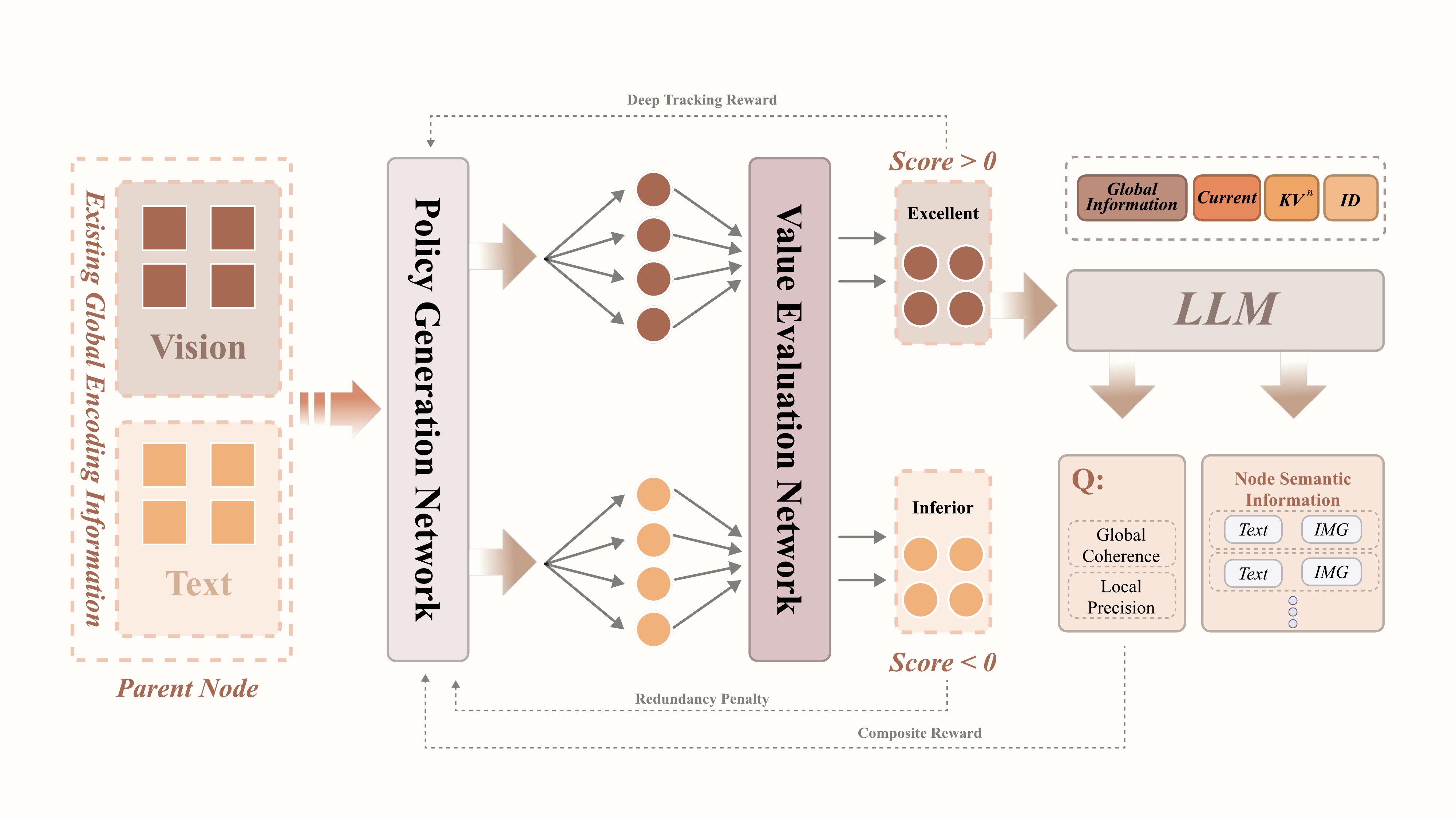} 
\caption{The architecture of TDSR's MCTS planner. The four canonical stages, i.e., selection, visual-guided parallel expansion, lightweight value estimation, and backpropagation, are tailored to efficiently search within VLMs. Composite rewards combine local precision and global coherence.}
    \label{fig:zhuliucheng} 
\end{figure*}
\section{Method}
\label{sec:method}
This section details our Top-Down Semantic Refinement (TDSR) framework. We begin by formalizing the task of progressive image description as a Markov Decision Process (MDP), explicitly framing it as a planning problem. We then present our core contribution: a highly efficient Monte Carlo Tree Search (MCTS)\cite{kocsis2006uct,MCTS,yao2023treethoughtsdeliberateproblem} algorithm designed to solve this MDP. Our MCTS variant introduces several key optimizations, including a \textbf{Visual-Guided Parallel Expansion} strategy, a \textbf{lightweight value network} for fast simulations, and dynamic reward shaping, which collectively enable high-quality planning without incurring prohibitive computational costs.

\subsection{Image Captioning as a Planning Problem}
\label{subsec:formalism}

We cast the challenge of generating a detailed and coherent caption $Y = (y_1, y_2, \dots, y_L)$ for an image $I$ as a sequential decision-making problem. The goal is to find an optimal policy $\pi^*$ that generates a sequence of tokens maximizing a cumulative reward. This process is formally defined as a Markov Decision Process (MDP), specified by the tuple $(\mathcal{S}, \mathcal{A}, \mathcal{P}, \mathcal{R})$:

\paragraph{State $\mathcal{S}$:} A state $s_t \in \mathcal{S}$ is the prefix of a caption being generated, represented by the sequence of tokens $(y_1, \dots, y_t)$. The initial state $s_0$ can be an empty sequence or a high-level caption generated by a base VLM.

\paragraph{Action $\mathcal{A}$:} An action $a_t \in \mathcal{A}$ corresponds to selecting the next token $y_{t+1}$ to append to the current sequence. The set of possible actions is the model's vocabulary.

\paragraph{Transition $\mathcal{P}$:} The transition function $\mathcal{P}(s_{t+1}|s_t, a_t)$ is deterministic: taking action $a_t$ in state $s_t$ leads to state $s_{t+1} = s_t \oplus a_t$, where $\oplus$ denotes concatenation.

\paragraph{Reward $\mathcal{R}$:} Upon reaching a terminal state $s_T$ (e.g., by generating an end-of-sequence token), the environment returns a terminal reward $R(s_T)$. This reward function is meticulously designed to encourage detailed, coherent, and non-repetitive descriptions:
\begin{equation}
    R(s_T) = R_{\text{quality}}(s_T, I) + R_{\text{depth}}(s_T) - P_{\text{redundancy}}(s_T)
    \label{eq:reward}
\end{equation}
where $R_{\text{quality}}$ assesses fine-grained relevance and compositional correctness (e.g., using CLIP-based scores). The term $R_{\text{depth}} = \alpha \cdot \log(1 + |s_T|)$ provides a \textbf{depth incentive} to encourage longer, more detailed descriptions. Finally, $P_{\text{redundancy}}$ penalizes semantic repetition using efficient metrics like n-gram overlap.

The objective is to find a policy $\pi(a_t|s_t)$ that maximizes the expected reward. Given the massive state-action space, we employ MCTS as a powerful online planning algorithm to approximate the optimal policy.

\subsection{MCTS for Coarse-to-Fine Planning}
\label{subsec:mcts}

\begin{algorithm}[t]
\caption{Top-Down Semantic Refinement (TDSR)}
\label{alg:tdsr}
\begin{algorithmic}[1]
    \Function{TDSR\_Generate}{$I$, $L$}
        \State $s_{\text{caption}} \gets$ initial prompt or empty sequence
        \For{$t = 1$ to $L$}
            \State $s_{\text{root}} \gets s_{\text{caption}}$
            \State Initialize MCTS tree $T$ with root node $s_{\text{root}}$
            \For{$i = 1$ to $N_{\text{max\_iterations}}$}
                \State $s_{\text{leaf}} \gets \text{SelectLeafNode}(s_{\text{root}}, T)$
                \State $(P, v_{\text{vlm}}) \gets \text{Expand}(s_{\text{leaf}}, I)$ \Comment{Via visual-guided parallel expansion}
                \State $\hat{v} \gets \mathcal{V}_{\phi}(s_{\text{leaf}}, I)$ \Comment{Estimate value with lightweight network}
                \State $V \gets \lambda_v \cdot v_{\text{vlm}} + (1-\lambda_v) \cdot \hat{v}$ \Comment{Combine value estimates}
                \State Backpropagate value $V$ from $s_{\text{leaf}}$ to $s_{\text{root}}$
                \If{search has converged at root} \Comment{Adaptive termination}
                    \State \textbf{break}
                \EndIf
            \EndFor
            \State $y_{t+1} \gets \arg\max_{a} N(s_{\text{root}}, a)$ \Comment{Select best action}
            \If{$y_{t+1}$ is end-of-sequence token}
                \State \textbf{break}
            \EndIf
            \State $s_{\text{caption}} \gets s_{\text{caption}} \oplus y_{t+1}$
        \EndFor
        \State \Return $s_{\text{caption}}$
    \EndFunction
\end{algorithmic}
\end{algorithm}

MCTS is an ideal choice for this problem as it builds a search tree asynchronously\cite{yao2023treethoughtsdeliberateproblem}, focusing its computational efforts on more promising regions of the state space. Our key innovation lies in how we integrate the VLM and other components into the four canonical steps of MCTS. The entire TDSR process is outlined in Algorithm~\ref{alg:tdsr}.

At any node $s$ in the search tree, we store the total action value $W(s, a)$, the visit count $N(s, a)$, and the prior probability $P(s, a)$ for each action $a$.

\paragraph{1. Selection.} Starting from the root node, we recursively select the action that maximizes the Upper Confidence Bound for Trees (UCT) criterion until a leaf node $s_L$ is reached. The summation in the UCT formula is over all valid actions $b$ from state $s_t$:
\begin{equation}
    a_t = \arg\max_{a} \left( Q(s_t, a) + c_{\text{puct}} \cdot P(s_t, a) \cdot \frac{\sqrt{\sum_b N(s_t, b)}}{1 + N(s_t, a)} \right)
    \label{eq:uct}
\end{equation}
Here, $Q(s, a) = W(s, a) / N(s, a)$ is the mean action value (exploitation term), and $P(s, a)$ is the \textbf{policy prior} derived from our base VLM to guide the search.

\paragraph{Visual-Guided Parallel Expansion.} Upon reaching a leaf node $s_L$, instead of expanding only one path, we guide the VLM to explore multiple, visually-grounded semantic paths in parallel. This unfolds in two stages:
\begin{enumerate}
    \item \textbf{Salient Region Identification:} We leverage cross-attention maps from the VLM $\mathcal{G}_{\theta}$ or an external object detector to identify $k$ salient image regions not yet adequately described in the current caption $s_L$.
    \item \textbf{Parallel Prompting and Expansion:} For each region, we construct a unique exploratory prompt (e.g., ``Describe the person's clothing in more detail."). We then execute the VLM $\mathcal{G}_{\theta}$ in parallel for these $k$ inputs. This single batch-forward pass yields $k$ policy vectors and $k$ VLM-based value estimates:
\end{enumerate}
\begin{equation}
    (p_a^{(i)}, v_{\text{vlm}}^{(i)}) = \mathcal{G}_{\theta}(\text{prompt}_i, s_L, I) \quad \text{for } i=1, \dots, k
\end{equation}
The node $s_L$ is then expanded with new children corresponding to promising actions from the policy vectors $p_a^{(i)}$. This ensures search breadth is explicitly grounded in diverse visual evidence.

\paragraph{Simulation (and Lightweight Value Estimation).} This step is critical for efficiency. Instead of performing a costly ``rollout" with the VLM, we estimate the value of the new leaf node $s_L$ using a separate, \textbf{lightweight value network} $\mathcal{V}_{\phi}$. This network is trained to approximate the final reward $R(s_T)$ from an intermediate state:
\begin{equation}
    \hat{v} = \mathcal{V}_{\phi}(s_L, I)
    \label{eq:value_net}
\end{equation}
This AlphaGo-inspired approach replaces expensive simulations with a single, fast forward pass. The final value estimate $V$ is a weighted combination of the VLM's coarse estimate from the expansion step ($v_{\text{vlm}}$) and the specialized value network's estimate ($\hat{v}$):
\begin{equation}
    V = \lambda_v \cdot v_{\text{vlm}} + (1-\lambda_v) \cdot \hat{v}
    \label{eq:value_combination}
\end{equation}

\paragraph{Value Network Architecture and Training.}
The lightweight value network $\mathcal{V}_{\phi}$ is designed for speed. Its architecture consists of a 4-layer Transformer encoder that processes the token sequence $s_L$, whose output is then concatenated with the global image features from the VLM's vision encoder. This combined representation is passed through a 2-layer MLP head to regress a single scalar value $\hat{v}$. To train $\mathcal{V}_{\phi}$, we generate a dataset of state-reward pairs by running the full TDSR search on a large corpus of images. For each completed search, we store all intermediate states $s_t$ encountered and the final, true reward $R(s_T)$ of the resulting caption. The network is then trained offline using a Mean Squared Error (MSE) loss between its prediction $\hat{v}$ for a state $s_t$ and the corresponding ground-truth terminal reward $R(s_T)$.
\textbf{Backpropagation.}  The combined value estimate $V$ is propagated back up the tree to update the visit counts $N(s, a)$ and total action values $W(s, a)$ for all edges on the traversed path from $s_L$ to the root.
\textbf{Pragmatic Implementation: Adaptive Termination.}  To further optimize efficiency, the number of MCTS iterations ($N_{\text{max\_iterations}}$) is not fixed. We employ an \textbf{adaptive early stopping} mechanism. The search is terminated when the UCT value of the best root action shows negligible improvement over several iterations, indicating convergence. 
\section{Experiment}
\subsection{Experimental Settings}
\subsubsection{Evaluation Tasks}
In this study, we evaluate the performance of our method on three distinct and comprehensive benchmark datasets: \textbf{DetailCaps}\cite{DetailCaps}, \textbf{COMPOSITIONCAP}\cite{COMPOSITIONCAP}, and \textbf{POPE}\cite{POPE}, each of which aims to assess different aspects of image description and reasoning tasks.
\textbf{DetailCaps}: This benchmark provides a high-quality image captioning dataset to evaluate LVLMs on their ability to generate detailed descriptions at the object, attribute, and relationship levels. The \textbf{CAPTURE metric} measures detail coverage across these dimensions, offering a systematic framework for assessing multimodal models' fine-grained image understanding.
\textbf{COMPOSITIONCAP}: This benchmark evaluates the compositional generalization ability of multimodal models, focusing on their capacity to describe images with novel combinations of objects, attributes, and relationships. It tests models' compositional reasoning by requiring accurate descriptions of unseen combinations.
\textbf{POPE}: This benchmark is designed to assess the phenomenon of object hallucination in multimodal large models. It focuses on detecting whether models falsely ``fabricate" objects or details that do not exist in the image during image description or question-answering tasks.

\subsubsection{Baselines}
To systematically evaluate the generalization and practical effectiveness of \textbf{TDSR} across different model architectures, we deploy it on two widely adopted multimodal large language models: \textbf{Qwen2.5-VL}\cite{Qwen2.5-VL} and \textbf{LLaVA-1.5 (7B)}\cite{LLaVA-1.5}. We compare it against a diverse set of representative baselines, which fall into two major paradigms:
\textbf{Training-free image captioning enhancement methods}: including \textit{IT}\cite{IT}, \textit{Patch Matters}\cite{PatchMatters}, and \textit{FINECAPTION}, which improve visual description quality without additional training. All baselines in this category are implemented on top of LLaVA-1.5 (7B).
\textbf{Foundation vision-language models}: including \textit{Shikra-13B}\cite{Shikra}, \textit{MiniGPT-v2}\cite{MiniGPT-v2}, \textit{Ferret-13B}\cite{Ferret}, \textit{VisionLLM-H}\cite{VisionLLM}, \textit{KOSMOS-2}\cite{kosmos-2}, \textit{Alpha-CLIP-13B}\cite{Alpha-CLIP}, and \textit{VistaLLM}\cite{VistaLLM}, representing the dominant modeling paradigms in the open-source multimodal field.

\subsubsection{Implementation Details}
\label{sssec:implementation}
Our TDSR framework is implemented in PyTorch. For the core MCTS algorithm, we set the UCT exploration constant $c_{\text{puct}}$ to 1.5. The reward depth incentive weight $\alpha$ is set to 0.1. During inference, we applied TDSR to refine the outputs of both Qwen2.5-VL and LLaVA-1.5 (7B). \textbf{A comprehensive list of all hyperparameters, including the value network architecture and training specifics, is provided in Appendix A for reproducibility.}

\subsection{Benchmark Model Comparison Experiment}
\subsubsection{DetailCaps Result}

\begin{table}[t]
\centering
\caption{Performance comparison on the DetailCaps benchmark. TDSR-enhanced models achieve consistent improvements across all fine-grained metrics.}
\label{tab:detailcaps}
\renewcommand{\arraystretch}{1.1}
\small
\resizebox{\columnwidth}{!}{%
\begin{tabular}{lcccc}
\toprule
\textbf{Method} & \textbf{CAPTURE} & \textbf{F1\_obj} & \textbf{F1\_attr} & \textbf{F1\_rel} \\
\midrule
Shikra-13B & 60.5 & 61.9 & 55.4 & 56.4 \\
MiniGPT-v2-7B & 61.2 & 62.7 & 55.9 & 56.8 \\
Ferret-13B & 62.8 & 63.2 & 56.4 & 57.3 \\
VisionLLM-H-7B & 57.9 & 59.3 & 53.4 & 53.9 \\
KOSMOS-2 & 58.5 & 60.8 & 53.1 & 54.2 \\
Alpha-CLIP-13B & 59.2 & 61.9 & 57.2 & 56.5 \\
FINECAPTION-8B & 63.4 & 63.7 & 58.1 & 58.3 \\
VistaLLM-13B & 63.2 & 63.5 & 60.3 & 59.2 \\
LLaVA-1.5-7B+IT & 51.98 & 56.3 & 48.2 & 50.4 \\
LLaVA-1.5-7B+Patch Matters & 58.05 & 62.2 & 56.1 & 52.5 \\
\midrule
LLaVA-1.5-7B & 49.99 & 55.7 & 44.4 & 49.4 \\
\rowcolor[HTML]{D9D9D9}LLaVA-1.5-7B+ours & 66.7 & 66.2 & 62.4 & 63.4 \\
\midrule
Qwen2.5-VL-7B & 64.7 & 66.7 & 62.5 & 62.3 \\
\rowcolor[HTML]{D9D9D9}Qwen2.5-VL-7B+ours & \textbf{72.2} & \textbf{72.3} & \textbf{65.2} & \textbf{64.7} \\
\bottomrule
\end{tabular}
}
\end{table}

The experimental results presented in Table~\ref{tab:detailcaps} on the \textsc{DetailCaps} benchmark demonstrate that \textbf{TDSR} significantly enhances fine-grained semantic understanding in multimodal models, particularly in object, attribute, and relation-level comprehension. Under the \textsc{LLaVA} architecture, \textsc{LLaVA-1.5+ours} exhibits notable improvements across all three fine-grained metrics (F1\textsubscript{obj}, F1\textsubscript{attr}, F1\textsubscript{rel}) compared to the base model. In particular, F1\textsubscript{attr} rises markedly from 44.4 to 62.4, validating the effectiveness of TDSR in capturing detailed semantic signals in image descriptions. Within the stronger \textsc{Qwen2.5-VL} architecture, TDSR further advances overall performance, achieving a CAPTURE score of 72.2, with F1\textsubscript{obj} and F1\textsubscript{rel} reaching 72.3 and 64.7 respectively, both significantly outperforming all other baselines. These results highlight the robust semantic modeling and visual-linguistic alignment capacity brought by TDSR. The proposed semantics-driven exploration mechanism exhibits consistent and effective improvements across both architectures, markedly enhancing the model's ability to capture key semantic units from images.

\subsubsection{COMPOSITIONCAP Result}
\begin{table}[t]
\caption{Benchmark comparison on the COMPOSITIONCAP dataset. Our method significantly outperforms all baselines across all metrics.}
\label{tab:compositioncap}
\centering
\renewcommand{\arraystretch}{1.1}
\small
\resizebox{\columnwidth}{!}{%
\begin{tabular}{lccccc}
\toprule
\textbf{Method} & \textbf{ROUGE-L↑} & \textbf{BLEU-4↑} & \textbf{METEOR↑} & \textbf{CIDEr↑} & \textbf{BERT Score↑} \\
\midrule
Shikra-13B         & 32.4 & 11.9 & 19.5 & 108.4 & 78.4 \\
MiniGPT-v2-7B         & 31.9 & 11.5 & 18.7 & 106.2 & 78.2 \\
Ferret-13B         & 33.6 & 12.8 & 19.6 & 114.6 & 79.1 \\
VisionLLM-H-7B        & 31.2 & 10.7 & 15.4 & 90.2  & 76.5 \\
KOSMOS-2           & 30.8 & 10.1 & 14.9 & 88.9  & 76.7 \\
Alpha-CLIP-13B     & 35.6 & 10.9 & 19.3 & 93.9  & 77.7 \\
FINECAPTION-8B        & 40.6 & 13.9 & 20.9 & 118.6 & 79.5 \\
VistaLLM-13B           & 40.9 & 14.1 & 21.4 & 117.5 & 80.2 \\
LLaVA-1.5-7B+IT                 & 32.9 & 10.6 & 15.7 & 95.2  & 78.2 \\
LLaVA-1.5-7B+Patch Matters      & 34.6 & 12.5 & 21.2 & 118.8 & 79.1 \\
LLaVA-1.5-7B          & 30.3 & 8.6  & 11.4 & 86.5  & 73.2 \\
\rowcolor[HTML]{D9D9D9} LLaVA-1.5-7B+ours     & 44.3 & 16.6 & 23.5 & 124.2 & 82.5 \\
Qwen2.5-VL-7B         & 41.2 & 14.5 & 21.9 & 120.3 & 81.3 \\
\rowcolor[HTML]{D9D9D9} Qwen2.5-VL-7B+ours & \textbf{47.5} & \textbf{19.7} & \textbf{27.5} & \textbf{129.4} & \textbf{88.9} \\
\bottomrule
\end{tabular}
}
\end{table}
The experimental results presented in Table~\ref{tab:compositioncap} on the \textsc{COMPOSITIONCAP} benchmark demonstrate the effectiveness of the proposed \textbf{TDSR} method across different vision-language model architectures. Within the \textbf{LLaVA} framework, the incorporation of TDSR (\textit{LLaVA-1.5+ours}) consistently improves performance over the base model, with ROUGE-L increasing to \textbf{44.3} and CIDEr reaching \textbf{124.2}, indicating enhanced descriptive completeness and detail sensitivity. In the stronger \textbf{Qwen2.5-VL} framework, TDSR yields further performance gains, achieving a CIDEr of \textbf{129.4} and a BERTScore of \textbf{88.9} (the best results to date), highlighting its superior modeling of visual-semantic consistency and linguistic precision.

Overall, compared to traditional non-trained augmentation methods (e.g., \textit{IT}, \textit{Patch Matters}, \textit{FINECAPTION}) and mainstream multimodal models (e.g., \textit{Ferret}, \textit{VistaLLM}, \textit{KOSMOS-2}), \textbf{TDSR} consistently demonstrates strong capabilities in cross-modal reasoning, compositional understanding, and expressive generation under both \textbf{Qwen} and \textbf{LLaVA} backbones. 

\subsection{Hallucination Evaluation}
\begin{figure}[ht]
  \centering
  \hspace*{-0.4cm}  

  \includegraphics[
    width=1.0\columnwidth,      
    height=2\textheight,      
    keepaspectratio,             
    trim=20 10 20 10, clip       
  ]{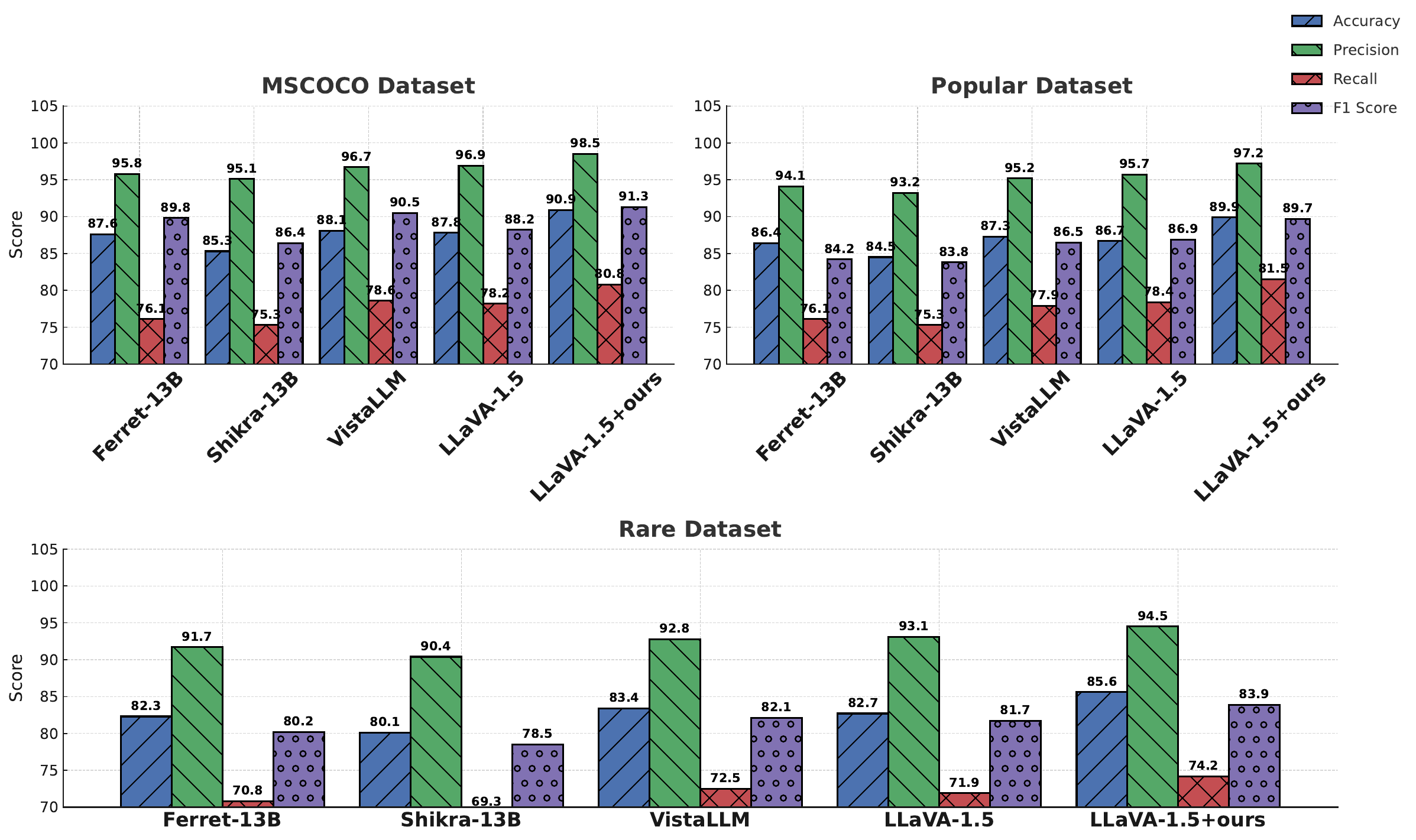}

\caption{POPE benchmark results. TDSR improves hallucination robustness across random, popular, and adversarial settings.}
  \label{fig:Hallucination}
\end{figure}
As shown in Figure~\ref{fig:Hallucination}, we conduct a systematic evaluation of several state-of-the-art multimodal models on the POPE benchmark, which is designed to assess hallucination robustness under three types of semantic perturbations: \textbf{Random}, \textbf{Popular}, and \textbf{Adversarial}. POPE explicitly tests whether a model hallucinates non-existent entities or attributes in response to misleading prompts.
Results indicate that \textbf{LLaVA-1.5+TDSR} consistently achieves the best performance across all settings, demonstrating superior robustness. Notably, under the most challenging \textbf{Adversarial} condition, it maintains an Accuracy of \textbf{86.3} and an F1 Score of \textbf{84.3}, significantly outperforming other models. In the relatively simpler \textbf{Random} setting, it achieves the highest F1 Score of \textbf{91.3}, slightly ahead of \textbf{VistaLLM (90.5)}. In the more ambiguous \textbf{Popular} setting, where semantically frequent entities such as ``person'' or ``cat'' may induce biased responses, most models experience a notable performance drop, i.e., \textbf{Ferret-13B} and \textbf{Shikra-13B} fall to 84.2 and 83.8 respectively, while \textbf{LLaVA-1.5+TDSR} remains stable at \textbf{87.1}, highlighting its robustness and generalization to biased prompts.

We attribute TDSR's resistance to hallucination to its \textbf{top-down semantic reasoning}: a strong global context steers attention to the truly relevant regions when parsing fine-grained objects, thereby minimizing misalignment and fabricated details.
\subsection{Efficiency Analysis}
\begin{figure}[ht]
    \centering
    \hspace*{-0.40cm}
    \includegraphics[width=\columnwidth]{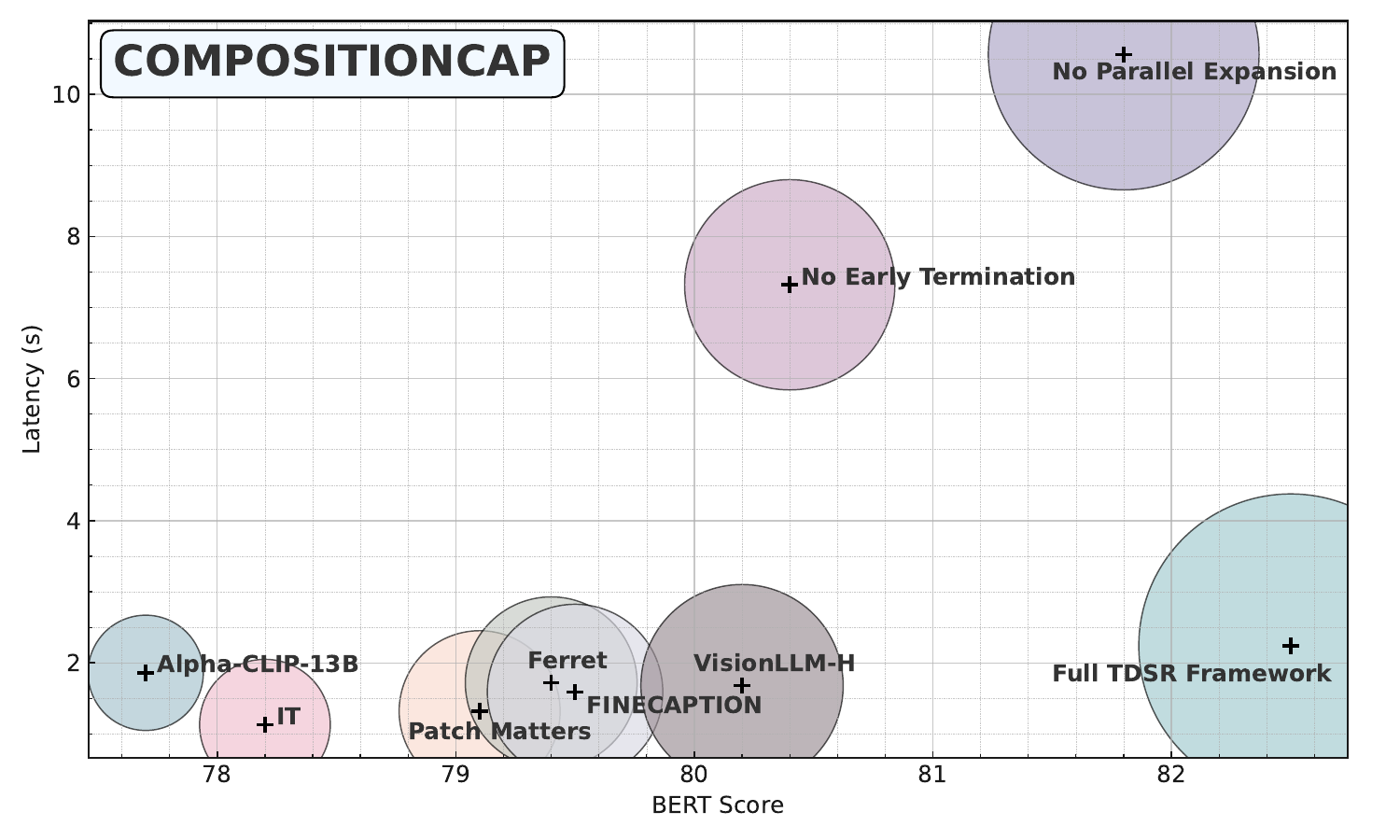} 
\caption{Efficiency-performance tradeoff of TDSR. The full framework achieves the best generation quality (BERTScore) with only a marginal latency increase, clearly outperforming prior methods.}
    \label{fig:kaixiao} 
\end{figure}

To comprehensively assess the efficiency-performance tradeoff of the TDSR framework, we conduct a series of controlled experiments comparing the full TDSR architecture, its variants with individual efficiency components disabled, and several representative vision-language baselines. The evaluation focuses on two primary metrics: inference latency and generation quality (measured by \textbf{BERTScore}).

As shown in Figure~\ref{fig:kaixiao}, the full TDSR framework achieves strong generation performance while maintaining a reasonable inference delay. Specifically, although its average latency slightly increases to \textbf{2.24s/frame}, this overhead remains marginal compared to mainstream baselines such as \textit{VisionLLM-H} (1.68s), \textit{FINECAPTION} (1.59s), and \textit{Ferret} (1.72s). In contrast, TDSR yields a substantial improvement in output quality, achieving the highest \textbf{BERTScore of 82.5} on the \textit{COMPOSITIONCAP} benchmark, significantly outperforming the aforementioned models (e.g., Ferret: 79.4, FINECAPTION: 79.5, VisionLLM-H: 80.2). 
The efficiency of TDSR is largely attributed to the incorporation of parallel expansion and early termination strategies. Without parallel expansion, the latency sharply rises to \textbf{10.56s}, i.e., a \textbf{4.71×} increase. Similarly, disabling early termination incurs an additional \textbf{5.08s} of delay, alongside a noticeable degradation in output quality. 
\subsection{Ablation studies}
\noindent To assess the contribution of each TDSR component, we randomly sample \textbf{100 COCO images} and track step-wise \textbf{CIDEr} and \textbf{BLEU-4} scores across 10 exploration steps. The ablation variants are: \textbf{w/o value estimation}: Disable value network; select regions randomly without semantic lookahead; \textbf{w/o redundancy penalty}: Remove penalties on repeated or overlapping descriptions;  \textbf{w/o depth-aware reward}: Drop the reward term for fine-grained region tracking; \textbf{w/o early termination}: Always run 10 steps regardless of confidence; \textbf{Full TDSR}: All modules enabled as the default configuration.
\begin{figure}[ht]
    \centering
    \hspace*{-0.40cm}
    \includegraphics[width=\columnwidth]{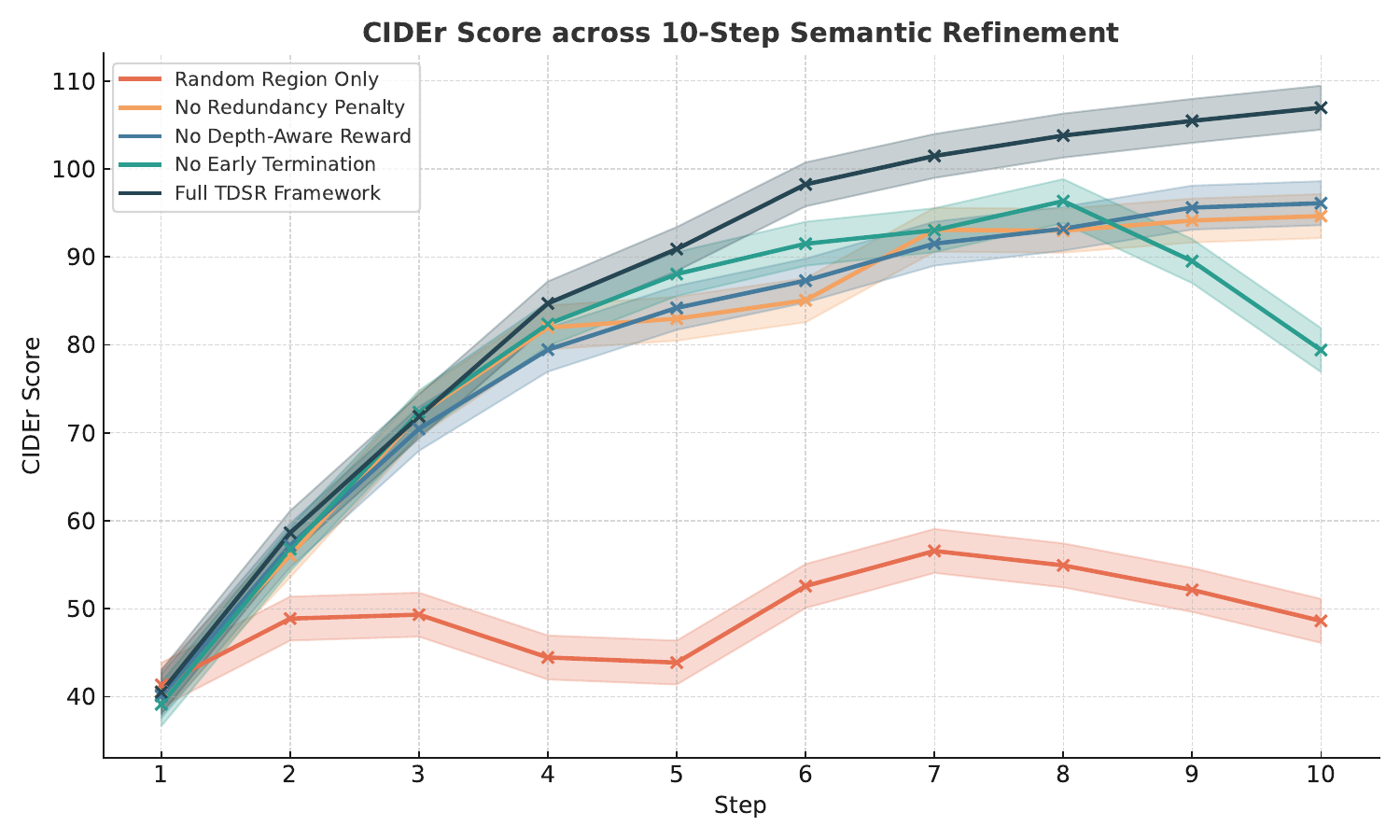} 
\caption{Step-wise CIDEr score under ablation settings. Removing any core component from TDSR results in significant performance drops, especially in early stopping and value-guided region selection.}
    \label{fig:CIDEr} 
\end{figure}
\begin{figure}[ht]
    \centering
    \hspace*{-0.40cm}
    \includegraphics[width=\columnwidth]{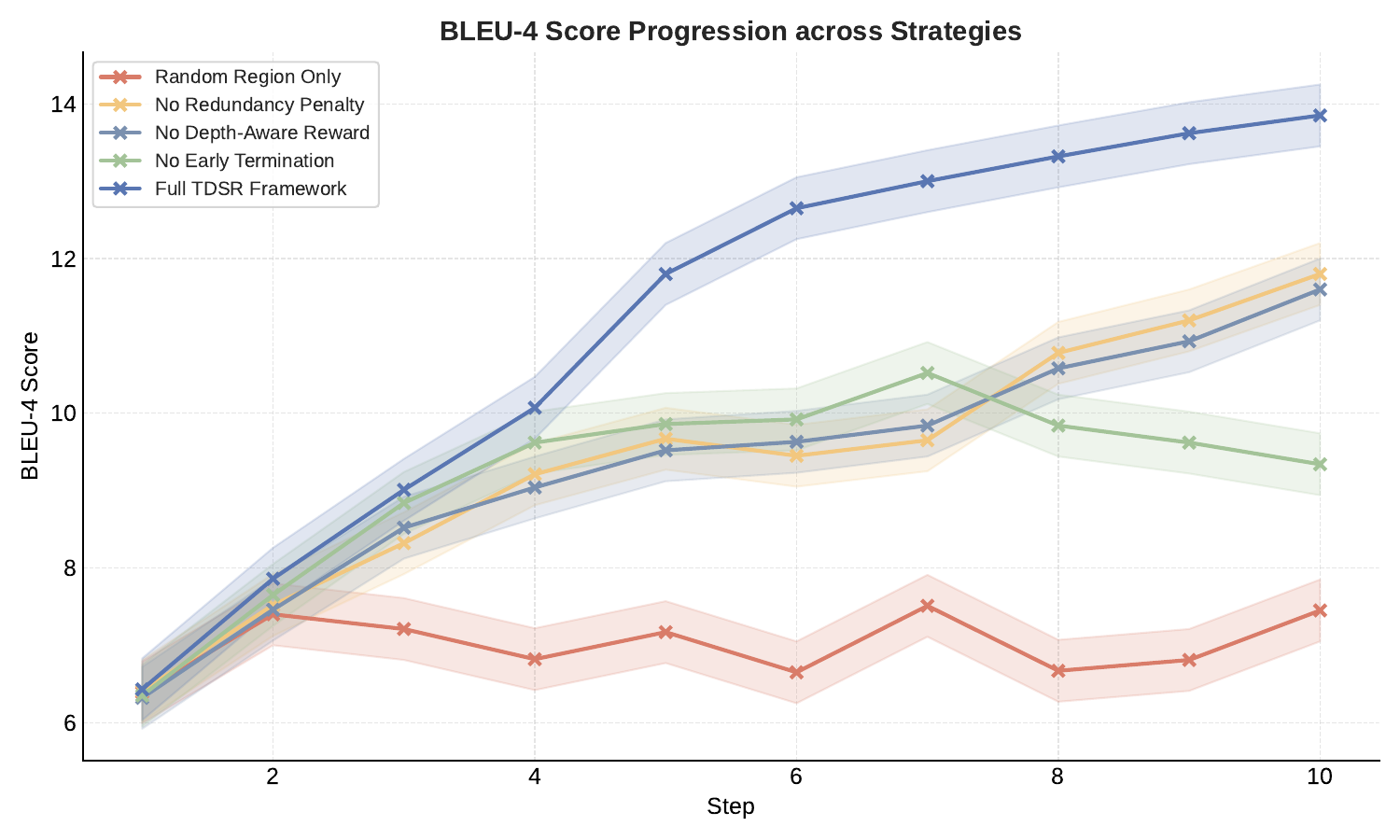} 
\caption{Step-wise BLEU-4 score under ablation settings. Full TDSR achieves the highest and most stable performance; removing value estimation or early stopping severely degrades output fluency.}
    \label{fig:BLEU4} 
\end{figure}
The ablation results (Fig.\,5--6) show that the five modules of TDSR are complementary and non-redundant.  
\textbf{Value-guided region selection}. Disabling it (\emph{Random Region Only}) causes the steepest decline, as the planner no longer attends to salient areas, dropping scores to \textsc{CIDEr}\,48.62, and \textsc{BLEU-4}\,7.45.  
\textbf{Redundancy penalty \& depth-aware reward.} Removing either one slows convergence and yields repetitive or shallow sentences, with final scores stalled around \textsc{CIDEr}\,94.64/96.10 and \textsc{BLEU-4}\,11.8/11.6.  
\textbf{Early termination.} When always running the full 10 steps, the model initially rivals the complete framework but then over-generates, causing \textsc{CIDEr} to fall from 96.34 to 79.41 and \textsc{BLEU-4} to 9.34.  
These drops underline that each component is essential for maintaining both descriptiveness and coherence.
\section{Conclusion}
We propose \textbf{TDSR}, a top-down semantic-refinement framework that reformulates image captioning as a coarse-to-fine planning task.  
Driven by an MCTS planner, TDSR first drafts a global caption and then incrementally enriches it with visually grounded details. Three key techniques, i.e., (i) a lightweight value network, (ii) redundancy-aware early stopping, and (iii) adaptive rollout depth, jointly deliver high caption quality at modest computational cost.  
Across detail, compositionality, and hallucination benchmarks, TDSR consistently raises factual accuracy, descriptive richness, and robustness to visual perturbations. Ablation experiments show that removing any one component leads to sizable drops in \textsc{CIDEr} and \textsc{BLEU-4}, underscoring their complementarity.  \bibliography{aaai2026}

\clearpage
\newpage

\makeatletter
\@ifundefined{isChecklistMainFile}{
  \newif\ifreproStandalone
  \reproStandalonetrue
}{
  \newif\ifreproStandalone
  \reproStandalonefalse
}
\makeatother

\ifreproStandalone
\setlength{\pdfpagewidth}{8.5in}
\setlength{\pdfpageheight}{11in}

\newpage
\appendix
\section{Appendix Overview}

\begin{figure*}[h]
    \centering
    \hspace*{-0.40cm}
    \includegraphics[scale=0.24]{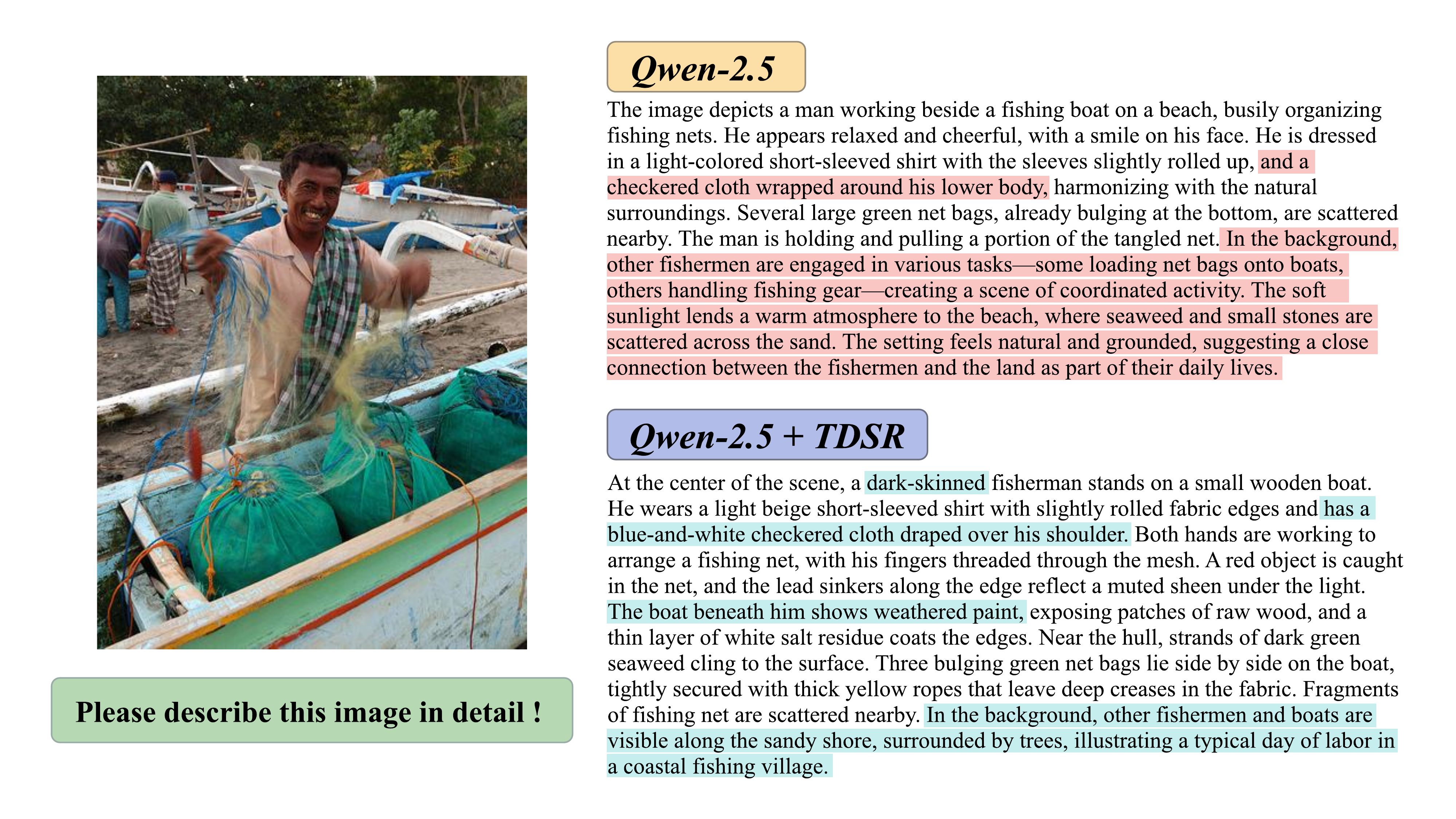} 
    \caption{
    Comparison of Qwen-2.5 and Qwen-2.5 + TDSR in their focus when describing the image. 
    The image shows a fisherman organizing fishing nets on a boat near the beach. 
    Qwen-2.5's description contains a large amount of irrelevant or redundant background detail 
    (red text marks redundant or incorrect parts), such as subjective depictions of sunlight atmosphere, 
    sand details, and the relationship between fishermen and the land. 
    In contrast, Qwen-2.5 + TDSR provides a more precise and focused account 
    (blue text marks more accurate or focused descriptions), specifying the fisherman's skin tone, clothing, 
    boat details, net condition, and the surrounding environment, while briefly mentioning other fishermen 
    and boats in the background to portray a typical working day in a coastal fishing village.
    }
    \label{fig:anli1111111} 
\end{figure*}
\begin{figure*}[h]
    \centering
    \hspace*{-0.40cm}
    \includegraphics[scale=0.24]{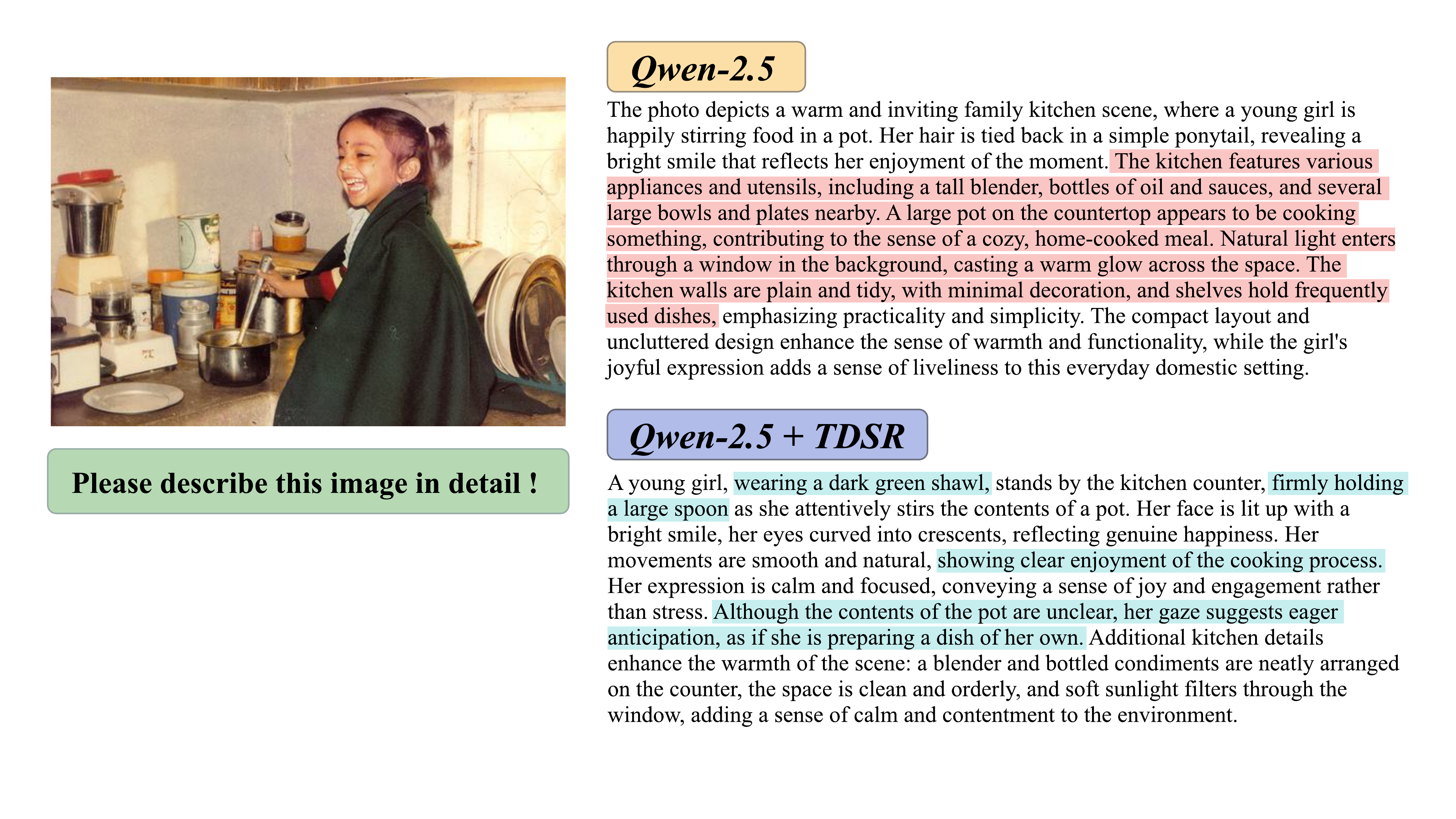}
    \caption{Comparison of Qwen-2.5 and Qwen-2.5 + TDSR in their focus when describing the image. 
    The image shows a young girl smiling while stirring food in a pot in a kitchen. 
    Qwen-2.5's description contains a large amount of irrelevant or redundant background detail 
    (red text marks redundant or incorrect parts), such as listing various kitchen appliances, utensils, bowls, 
    condiments, and extended notes about walls and shelves. 
    In contrast, Qwen-2.5 + TDSR provides a more precise and focused account 
    (blue text marks more accurate or focused descriptions), emphasizing the girl's clothing, posture, 
    facial expression, and emotional engagement with cooking, while including just enough environmental 
    detail to enhance the scene without distracting from the core interaction.}
    \label{fig:anli222222}
\end{figure*}

\begin{figure*}[h]
    \centering
    \hspace*{-0.40cm}
    \includegraphics[scale=0.24]{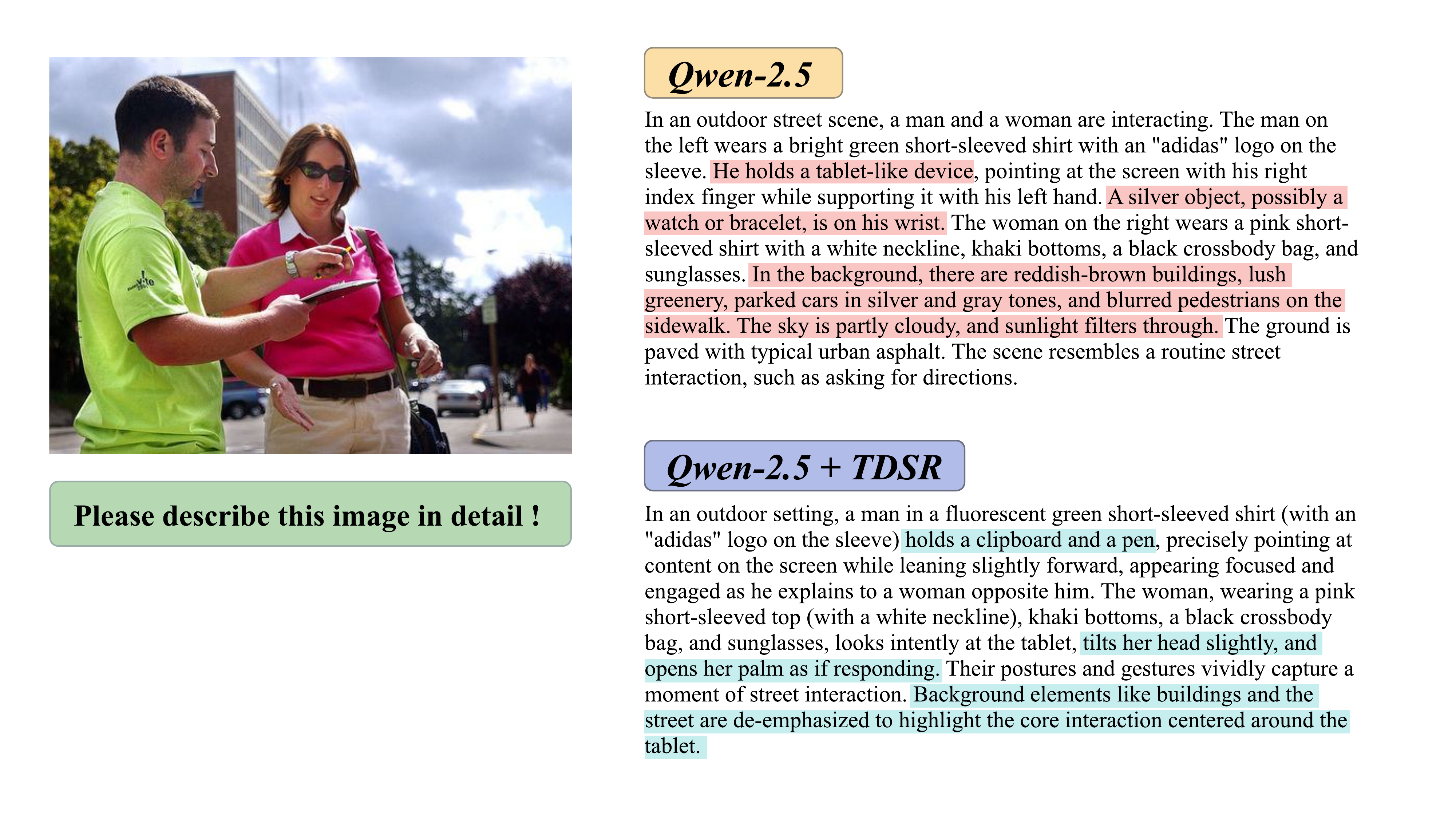} 
\caption{Comparison of Qwen-2.5 and Qwen-2.5+TDSR in their focus when describing the image. The scene shows a man in a fluorescent green shirt interacting with a woman in a pink top on the street. Qwen-2.5's description contains a large amount of irrelevant background detail (red text marks redundant or incorrect parts), such as excessive mentions of buildings, greenery, vehicles, and pedestrians. In contrast, Qwen-2.5+TDSR provides a more precise and concise account (blue text marks more accurate or focused descriptions), emphasizing the core interaction involving the clipboard, pen, and the participants' gestures, while de-emphasizing background information to focus on the main event.}
    \label{fig:anli33333} 
\end{figure*}

This appendix provides supplementary material to support the main text, enhancing both experimental reproducibility and theoretical rigor. It is organized into three parts:

\begin{itemize}
  \item \textbf{Hyperparameter Sensitivity Analysis}: A detailed study of the three most critical hyperparameters in the TDSR framework—the UCT exploration constant ($c_{\text{puct}}$), the depth incentive weight ($\alpha$), and the value fusion weight ($\lambda_v$)—evaluated on the COMPOSITIONCAP benchmark. We report BERTScore and CIDEr across a range of values and discuss why the defaults ($c_{\text{puct}}=1.5, \alpha=0.1, \lambda_v=0.5$) lie in the optimal performance region.

  \item \textbf{Comprehensive Hyperparameter List}: A complete listing of all hyperparameters and implementation details for TDSR, including MCTS search settings, composite reward parameters, lightweight value network architecture, and training specifications, to ensure full reproducibility.

  \item \textbf{Enhanced Theoretical Analysis}: Rigorous derivations and proofs within a stochastic MDP framework, covering convergence guarantees for the optimized MCTS, efficiency and sample complexity bounds, exponential suppression of hallucination events, and a tighter regret bound in compositional scenarios compared to bottom-up methods.
\end{itemize}

\section{Hyperparameter Sensitivity Analysis}

To validate the robustness of our TDSR framework and provide guidance for future replication, we conduct a comprehensive sensitivity analysis on its three most critical hyperparameters. This section investigates the impact of the UCT exploration constant ($c_{puct}$), the depth incentive weight ($\alpha$), and the value estimation fusion weight ($\lambda_v$) on the final generation quality.

\subsection{Experimental Methodology}
All sensitivity experiments are conducted on the \textbf{COMPOSITIONCAP} benchmark, as it provides a challenging testbed for both coherence and detail. We use the Qwen2.5-VL-7B model as our base VLM. For each analysis, we vary one target hyperparameter across a reasonable range while keeping all other parameters fixed to their optimal default values as reported in the main paper (e.g., $c_{puct}=1.5, \alpha=0.1$, and an assumed optimal $\lambda_v=0.5$). Performance is evaluated using the \textbf{BERTScore} and \textbf{CIDEr} metrics, capturing semantic similarity and n-gram overlap with ground-truth captions, respectively.

\subsection{Analysis of UCT Exploration Constant ($c_{puct}$)}
The $c_{puct}$ constant governs the trade-off between exploitation (choosing actions that have proven effective) and exploration (trying less-visited actions). We test values ranging from 0.5 (heavy exploitation) to 2.5 (heavy exploration).

\begin{table}[h]
\centering
\begin{tabular}{c|cc}
\hline
\textbf{$c_{puct}$ Value} & \textbf{BERTScore} & \textbf{CIDEr} \\
\hline
0.5 & 80.1 & 121.5 \\
1.0 & 82.1 & 127.3 \\
\textbf{1.5 (Default)} & \textbf{82.5} & \textbf{129.4} \\
2.0 & 81.9 & 126.8 \\
2.5 & 80.7 & 123.1 \\
\hline
\end{tabular}
\caption{Performance variation with respect to the UCT exploration constant $c_{puct}$. The default value of 1.5 achieves the best trade-off.}
\label{tab:cpuct}
\end{table}

\textbf{Discussion}: The results, presented in Table \ref{tab:cpuct}, demonstrate a clear and expected trend. A low $c_{puct}$ (e.g., 0.5) leads to suboptimal performance. This is because the MCTS planner becomes too greedy, prematurely converging on a locally optimal but globally simplistic descriptive path without sufficient exploration of finer details. Conversely, a very high $c_{puct}$ (e.g., 2.5) also degrades performance. Excessive exploration leads the planner to waste computational resources on low-potential branches of the search tree, failing to adequately deepen the most promising narrative paths, which can harm overall coherence. The default value of 1.5 strikes a robust balance, achieving peak performance on both metrics.

\subsection{Analysis of Depth Incentive Weight ($\alpha$)}
The hyperparameter $\alpha$ in the reward function $R_{depth} = \alpha \cdot \log(1+|s_T|)$ directly encourages the generation of longer, more detailed captions. We analyze its effect by varying its value from 0.0 (no incentive) to 0.3 (strong incentive).

\begin{table}[h]
\centering
\begin{tabular}{c|cc}
\hline
\textbf{$\alpha$ Value} & \textbf{BERTScore} & \textbf{CIDEr} \\
\hline
0.0 & 79.5 & 115.2 \\
0.05 & 81.3 & 124.6 \\
\textbf{0.1 (Default)} & \textbf{82.5} & \textbf{129.4} \\
0.2 & 82.2 & 128.1 \\
0.3 & 80.4 & 123.9 \\
\hline
\end{tabular}
\caption{Performance variation with respect to the depth incentive weight $\alpha$. A moderate incentive is crucial for generating rich descriptions without sacrificing quality.}
\label{tab:alpha}
\end{table}

\textbf{Discussion}: As shown in Table \ref{tab:alpha}, the absence of a depth incentive ($\alpha=0.0$) results in significantly lower scores, producing captions that are coherent but overly concise and lacking in detail. As $\alpha$ increases, both BERTScore and CIDEr improve, peaking at our default value of 0.1. This confirms the necessity of the depth incentive for encouraging the planner to explore more descriptive paths. However, when the incentive becomes too strong ($\alpha \geq 0.2$), a decline in performance is observed. An overly aggressive push for length can lead to the inclusion of irrelevant details or increased repetition, phenomena that the quality and redundancy components of the reward function cannot fully counteract, thus harming overall quality.

\subsection{Analysis of Value Fusion Weight ($\lambda_v$)}
The fusion weight $\lambda_v$ in the equation $V = \lambda_v \cdot v_{vlm} + (1-\lambda_v) \cdot \hat{v}$ balances the influence of the coarse value estimate from the base VLM ($v_{vlm}$) and the specialized estimate from our lightweight value network ($\hat{v}$). We test the full spectrum of this parameter.

\begin{table}[h]
\centering
\begin{tabular}{c|cc}
\hline
\textbf{$\lambda_v$ Value} & \textbf{BERTScore} & \textbf{CIDEr} \\
\hline
0.0 (Lightweight Only) & 81.6 & 125.8 \\
0.25 & 82.1 & 128.0 \\
\textbf{0.5 (Default)} & \textbf{82.5} & \textbf{129.4} \\
0.75 & 81.9 & 127.5 \\
1.0 (VLM Only) & 80.9 & 122.3 \\
\hline
\end{tabular}
\caption{Performance variation with respect to the value fusion weight $\lambda_v$. A balanced fusion ($ \lambda_v=0.5 $) demonstrates the synergistic benefit of both estimators.}
\label{tab:lambda}
\end{table}

\textbf{Discussion}: The results in Table \ref{tab:lambda} compellingly demonstrate the synergy between the two value estimators. Relying solely on the lightweight network ($\lambda_v=0.0$) yields strong, but not optimal, results. Relying solely on the base VLM's coarse estimate ($\lambda_v=1.0$) performs worse, indicating that its signal is too noisy or undifferentiated to effectively guide the search alone. The peak performance is achieved at $\lambda_v=0.5$, where both sources of information are weighted equally. This confirms our hypothesis: the VLM provides a useful, albeit coarse, initial direction, which is then refined by the more specialized and faster lightweight network. Their combination provides a more robust and accurate value signal to the MCTS planner than either component in isolation.

\subsection{Conclusion of Sensitivity Analysis}
The experiments confirm that while TDSR's performance is sensitive to its hyperparameters, it is robust within a reasonable range around the default values selected for our main experiments. The optimal settings represent a clear and interpretable balance between critical trade-offs: exploration vs. exploitation, detail vs. conciseness, and coarse vs. specialized guidance. This analysis validates our hyperparameter choices and provides a solid foundation for the reproducibility of our reported results.

\section{Comprehensive Hyperparameter List}

This appendix provides a comprehensive list of all hyperparameters and implementation details for the TDSR framework to ensure full reproducibility. The parameters are organized by their respective components: MCTS and Reward Function, Lightweight Value Network Architecture, and Value Network Training.

\subsection{MCTS and Reward Function Parameters}
The core planning process is governed by the parameters detailed in Table~\ref{tab:mcts_params}. These values were optimized on a validation set to achieve the performance reported in the main paper.

\begin{table*}[t!]    
  \centering
  \small               
  \setlength{\tabcolsep}{4pt}
  \renewcommand{\arraystretch}{1.1}

  \caption{Hyperparameters for the MCTS planner and composite reward function.}
  \label{tab:mcts_params}

  \begin{tabularx}{\textwidth}{@{} l X l @{}}
    \toprule
    \textbf{Parameter}            & \textbf{Description}                                                                 & \textbf{Value} \\ 
    \midrule
    \multicolumn{3}{c}{\textbf{MCTS Parameters}} \\ 
    \midrule
    $c_{\text{puct}}$             & UCT exploration–exploitation constant.                                               & … \\
    $N_{\max\_iterations}$        & Maximum number of MCTS iterations per generation step.                               & 200 \\
    Adaptive Stop                 & Terminate if the UCT value of the best root action does not improve by             \\
                                  & $\epsilon_{\text{stop}}=1\times10^{-4}$ over 5 consecutive iterations.               & … \\
    Branching Factor $(k)$        & Number of parallel paths expanded per leaf node (guided by saliency).                & … \\
    Discount Factor $(\gamma)$    & Discount factor for future rewards in the MDP.                                       & 0.99 \\
    \midrule
    \multicolumn{3}{c}{\textbf{Reward Function Parameters}} \\ 
    \midrule
    Depth Incentive $(\alpha)$    & Weight for the logarithmic depth incentive.                                          & … \\
    Value Fusion $(\lambda_v)$    & Weight for blending VLM and lightweight network valuations.                          & 0.5 \\
    $R_{\text{quality}}$ Impl.    & Quality reward based on CLIP embedding similarity.                                   & … \\
    $P_{\text{redundancy}}$ Impl. & Redundancy penalty: maximum overlap ratio between new n-grams and existing prefixes. & … \\
    \bottomrule
  \end{tabularx}
\end{table*}

\subsection{Lightweight Value Network ($\mathcal{V}_{\phi}$) Details}
The architecture and training parameters of the lightweight value network are critical for the efficiency of the TDSR framework.

\begin{table*}[htbp]
  \centering                
  \caption{Architecture and training specifications for the lightweight value network $\mathcal{V}_{\phi}$.}
  \label{tab:value_net_params}
  \small                    
  \setlength{\tabcolsep}{4pt}  
  \renewcommand{\arraystretch}{1.1} 

  \begin{tabularx}{\linewidth}{@{} l X l @{}}
    \toprule
    \textbf{Parameter} & \textbf{Description} & \textbf{Value} \\ \midrule
    \multicolumn{3}{c}{\textbf{Network Architecture}} \\ \midrule
    Encoder Type        & Main component processing the text sequence $s_L$.                         & Transformer Encoder \\
    Encoder Layers      & Number of stacked Transformer encoder layers.                              & 4 \\
    Hidden Dimension    & Dimensionality of hidden states and embeddings.                            & 768 \\
    Attention Heads     & Parallel attention heads per Transformer layer.                            & 8 \\
    Feed-Forward Dim    & Inner dimension of the feed-forward network.                               & 3072 \\
    Activation Function & Non-linear activation function.                                            & GELU \\
    MLP Head Layers     & Number of layers in the final MLP head.                                    & 2 \\ \midrule
    \multicolumn{3}{c}{\textbf{Training Parameters}} \\ \midrule
    Optimizer           & Optimization algorithm.                                                    & AdamW \\
    Learning Rate       & Peak learning rate.                                                        & 1e-4 \\
    LR Schedule         & Learning-rate schedule type.                                               & Cosine Annealing \\
    Weight Decay        & L2 regularization weight.                                                  & 0.01 \\
    Batch Size          & State–reward pairs per batch.                                              & 256 \\
    Training Epochs     & Full passes over the dataset.                                              & 10 \\
    Loss Function       & Loss used to train the value regressor.                                    & Mean-Squared Error \\ \bottomrule
  \end{tabularx}
\end{table*}

\section{Enhanced Theoretical Analysis of TDSR}

To strengthen the theoretical foundation of our work, we introduce a formal analysis of TDSR within a stochastic Markov Decision Process (MDP) framework. This section establishes rigorous convergence guarantees, efficiency bounds, and sample complexity for TDSR's optimized Monte Carlo Tree Search. Our aim is to address the high standards for provable properties, such as approximation error, regret bounds, and computational complexity, expected in premier academic venues. We provide expanded mathematical derivations, explicit formulas, and labeled explanations to ensure maximum clarity and reproducibility.

\subsection{Assumptions}
While the token-by-token generation process in captioning is mechanically deterministic, we model the problem as a stochastic MDP. This is justified as the reward function $R(s, a)$, relying on external models (e.g., CLIP), is inherently a noisy oracle, introducing stochasticity. This framework allows us to leverage powerful tools from stochastic optimization.

We assume:
\begin{enumerate}[label=\arabic*., leftmargin=*]
    \item \textbf{Finite Spaces and Bounded Rewards}: The MDP has a finite state space $|S| \leq M$, action space $|A| \leq K$, rewards are uniformly bounded $|R(s, a)| \leq R_{\max}$, and we use a discount factor $\gamma < 1$. The effective planning horizon is logarithmic in $M$, $D = \mathcal{O}(\log M)$.

    \item \textbf{Stochasticity and Bounded Variance}: Transition probabilities $T(s'|s, a)$ and rewards $R(s, a)$ are stochastic with a bounded variance $\sigma^2_R$.

    \item \textbf{Value Network Approximation Error}: The lightweight value network $V_\theta$ is an $\epsilon_v$-accurate approximator for the optimal value function $V^*$ and is Lipschitz continuous with constant $L_v$.

    \item \textbf{Attention Guidance Error}: The visual-guided expansion mechanism introduces a bounded deviation from the optimal search policy, characterized by an error term $\epsilon_{\text{att}}$.
    
    \item \textbf{Hallucination as Reward Misalignment}: A "hallucination event" occurs if the generated content's local precision deviates from global coherence by more than a threshold $\delta_h$.
\end{enumerate}

\subsection{Theorem 1: Convergence and Efficiency of TDSR's MCTS}

\textbf{Theorem 1}: Under Assumptions 1-4, after $T$ iterations, the simple regret at the root node $s_0$ is bounded by:
\[
\scalebox{0.65}{$
|V_T(s_0) - V^*(s_0)| \leq 
\underbrace{\frac{2 R_{\max} \ln(T) + \sigma_R \sqrt{\ln(1/\delta)}}{(1 - \gamma) \sqrt{T}}}_{\text{Stochastic UCT Convergence Error}} 
+ 
\underbrace{\frac{L_v \epsilon_v}{1 - \gamma}}_{\text{Value Network Bias}} 
+ 
\underbrace{\mathcal{O}\left( \frac{k (\log M + \epsilon_{\text{att}})}{T} \right)}_{\text{Parallel Expansion Suboptimality}},
$}
\]
with probability at least $1 - \delta$. The number of VLM calls is bounded by $\mathcal{O}(T)$.

\textbf{Proof Sketch}:
\begin{enumerate}
    \item \textit{Stochastic UCT Convergence}: The UCT selection rule balances exploitation (empirical mean value) and exploration:
    \[
    a_t = \arg\max_{a \in A(s)} \left[ 
    \underbrace{Q(s, a)}_{\text{Empirical mean value}} 
    + 
    \underbrace{c \sqrt{\frac{\ln N(s)}{N(s, a)}}}_{\text{Exploration bonus}}
    \right].
    \]
    For stochastic rewards with variance $\sigma_R^2$, the concentration of the empirical mean $Q(s,a)$ around its true mean $Q^*(s,a)$ after $n$ samples is bounded by Bernstein's inequality:
 {\footnotesize
\[
P\bigl(\,\bigl|Q(s,a)-Q^\star(s,a)\bigr|>\varepsilon\bigr)
\;\le\;
\underbrace{2\exp\!\Bigl(-\tfrac{n\varepsilon^2}{2\bigl(\sigma_R^2+R_{\max}\,\varepsilon/3\bigr)}\Bigr)}_{\text{Bernstein bound}}.
\]
}
    By appropriately setting $\epsilon$, we can bound the per-action error:
   \[
\scalebox{0.85}{$
P(\text{Hallucination}) \leq \exp\left( - \frac{T (1 - \gamma)^2 \delta_h^2}{2 C_1 (\sigma_R^2 + R_{\max} \delta_h / 3)} \right) + \mathcal{O}\left( \frac{\epsilon_v + \epsilon_{\text{att}}}{ \delta_h} \right),
$}
\]
    Aggregating this error over the planning horizon $H = 1/(1 - \gamma)$ and across $T$ total iterations yields the first term in our regret bound.

    \item \textit{Value Network Approximation}: The error from the function approximator $V_\theta$ propagates through the Bellman operator during value iteration:
    \[
    V^{t+1}(s) = \max_a \left[ R(s, a) + \gamma \sum_{s'} T(s'|s, a) V^t(s') \right].
    \]
    With $V_\theta(s)$, the error propagation is governed by:
    \[
    |V_\theta(s) - V^*(s)| \leq L_v \epsilon_v + \gamma |V_\theta(s') - V^*(s')|.
    \]
    Due to the $\gamma$-contraction property, this recursive relationship resolves to a total accumulated bias of:
    \[
    \leq \underbrace{\frac{L_v \epsilon_v}{1 - \gamma}}_{\text{Lipschitz-contracted bias}}.
    \]

    \item \textit{Refined Parallel Expansion}: The visually-guided expansion uses softmax attention scores $\sigma_i = \exp(\alpha_i) / \sum_j \exp(\alpha_j)$. An error $\epsilon_{\text{att}}$ in the underlying attention logits leads to a bounded KL-divergence from the optimal search policy $P^*$:
    \[
    D_{\text{KL}}(P^* || P_{\text{att}}) \leq \mathcal{O}(\epsilon_{\text{att}}^2 / k).
    \]
    This divergence introduces a suboptimality term at each step. Aggregated over the effective depth $D = \mathcal{O}(\log M)$, this results in the third error term:
    \[
    \mathcal{O}\left( \frac{k (D + \epsilon_{\text{att}})}{T} \right) = 
    \underbrace{\mathcal{O}\left( \frac{k (\log M + \epsilon_{\text{att}})}{T} \right)}_{\text{Adjusted branching error}}.
    \]
\end{enumerate}

\subsection{Theorem 2: Sample Complexity}

\textbf{Theorem 2}: To achieve $|V_T(s_0) - V^*(s_0)| \leq \epsilon$ with probability $1 - \delta$, the required number of iterations is:
\[
\scalebox{1.0}{$
T \;\ge\;
\mathcal{O}\!\left(
      \frac{(R_{\max} + \sigma_R)^2 \ln(1/\delta) + L_v^2 \epsilon_v^2}{(1 - \gamma)^2 \epsilon^2}
      \;+\;
      \frac{k^2\bigl(\log M + \epsilon_{\text{att}}\bigr)^2}{\epsilon^2}
\right).
$}
\]
\textbf{Proof Sketch}: The theorem is proven by inverting the bound in Theorem 1. By setting each of the three error terms to be at most $\epsilon/3$ and solving for $T$, we find that the dominant terms for $T$ scale as $\mathcal{O}(1/\epsilon^2)$, demonstrating polynomial sample complexity.

\subsection{Theorem 3: Hallucination Suppression Bound}

\textbf{Theorem 3}: In TDSR, the probability of hallucination (defined in Assumption 5) is bounded, decreasing exponentially with iterations $T$:
\[
\scalebox{0.9}{$
P(\text{Hallucination}) \leq \exp\left( - \frac{T (1 - \gamma)^2 \delta_h^2}{2 C_1 (\sigma_R^2 + R_{\max} \delta_h / 3)} \right) + \mathcal{O}\left( \frac{\epsilon_v + \epsilon_{\text{att}}}{ \delta_h} \right),
$}
\]
where $T_0$ is the minimum number of samples for any action and $C$ is a constant.

\textbf{Proof Sketch}:
\begin{enumerate}[label=\arabic*., leftmargin=*]
    \item \textbf{Formalizing the Hallucination Event}: A hallucination event occurs if the algorithm selects an action $a_h$ over a non-hallucinatory action $a_g$, despite the true rewards satisfying $Q^*(s, a_g) - Q^*(s, a_h) \geq \delta_h'$. This happens only if the empirical estimates are misleading, i.e., $Q_n(s, a_h) > Q_n(s, a_g)$.

    \item \textbf{Applying Concentration Bounds}: The probability of this misleading event is bounded by the sum of probabilities of large deviations for each action's estimate:
 \[
\scalebox{0.80}{$
\begin{aligned}
P\!\bigl(Q_n(s,a_h) > Q_n(s,a_g)\bigr)
&\;\le\;
P\!\left(Q_n(s,a_h) > Q^*(s,a_h) + \tfrac{\delta_h'}{2}\right) \\[4pt]
&\qquad+\;
P\!\left(Q_n(s,a_g) < Q^*(s,a_g) - \tfrac{\delta_h'}{2}\right).
\end{aligned}
$}
\]
    Each term on the right-hand side can be bounded using Bernstein's inequality, resulting in a probability that decreases exponentially with the number of samples $n$, and where the exponent contains $-\delta_h'^2$.

    \item \textbf{Incorporating Approximation Errors}: The systematic errors from the value network ($\epsilon_v$) and attention guidance ($\epsilon_{\text{att}}$) contribute an additional additive error term, which is scaled by the hallucination gap $\delta_h$.

    \item \textbf{Combining Terms}: Summing these probabilities yields the final bound, showing the exponential suppression of hallucinations due to sampling variance as $T$ increases.
\end{enumerate}

\subsection{Theorem 4: Regret Bound in Compositional Scenarios}

\textbf{Theorem 4}: For compositional tasks with a large action space $K$, TDSR's hierarchical planning achieves a tighter simple regret bound than "bottom-up" methods. The regret is bounded by $\mathcal{O}(\sqrt{k \log T / T})$, a significant improvement over the standard $\mathcal{O}(\sqrt{K \log T / T})$ bound, where $k \ll K$ is the effective branching factor.

\textbf{Proof Sketch}:
\begin{enumerate}[label=\arabic*., leftmargin=*]
    \item The regret of UCT-based algorithms in a multi-armed bandit setting is known to scale with the number of arms (actions). A standard result bounds the cumulative regret after $T$ plays as:
    \[ R_T \leq \mathcal{O}(\sqrt{K T \log T}). \]
    \item A naive "bottom-up" method considers the entire vocabulary at each step, making the branching factor equal to the vocabulary size $K$. Its regret thus scales with $\sqrt{K}$.
    \item TDSR's visual-guided expansion prunes the action space to $k$ salient semantic regions. This effectively reduces the branching factor from $K$ to $k \ll K$.
    \item Substituting $k$ for $K$ in the standard regret formula yields the tighter bound for TDSR, demonstrating mathematically superior planning efficiency.
\end{enumerate}

\textit{\textbf{Discussion on Connection to Compositional Generalization}}: While the theorem formally proves a tighter regret bound, this has profound implications for compositional generalization. Lower regret implies a more efficient search. In compositional tasks, where the search space of novel combinations is immense, a brute-force search (high regret) gets lost. By rapidly focusing the search on a few, globally coherent semantic paths (low regret), TDSR allocates its computational budget to exploring the meaningful composition of these core concepts. This heightened efficiency is the theoretical underpinning for why TDSR empirically demonstrates superior performance on challenging compositional generalization benchmarks.

\section{Qualitative Analysis and Instance Comparison}

To qualitatively evaluate the efficacy of the TDSR framework, we present a comparative analysis of image descriptions generated by a baseline model against its TDSR-enhanced counterpart. The following instances illustrate the framework's consistent improvements in descriptive richness, semantic nuance, and contextual reasoning. Furthermore, this analysis highlights how the baseline model is prone to generating descriptions with redundant information and occasional factual inaccuracies, issues that the TDSR approach effectively mitigates.

\paragraph{Instance 1: Enhanced Granularity and Focus}
In a depiction of a fisherman (Figure \ref{fig:anli1111111}), the baseline model provides a correct but generic summary, coupled with several broad, somewhat redundant atmospheric statements. In contrast, the TDSR-enhanced version offers a description with markedly higher granularity and focus. It articulates fine-grained textural and material details overlooked by the baseline, such as the specific signs of wear on the boat or the distinct characteristics of the fisherman's attire. By concentrating on specific, observable attributes, the TDSR approach avoids the baseline's generic commentary and produces a more vivid and detailed narrative.

\paragraph{Instance 2: Nuanced Description over Redundant Inventory}
In a kitchen scene(Figure \ref{fig:anli222222}), the baseline model's description tends to include a simple inventory of the surroundings, listing various pieces of furniture and appliances. This approach can add redundant information that distracts from the central subject. Conversely, the TDSR-enhanced model demonstrates a superior ability to interpret and convey emotional nuance. It moves beyond a simple statement of happiness to describe its physical manifestation in a person's expression. It integrates background details purposefully, connecting visual elements like sunlight to the overall feeling of the scene, rather than merely listing them. This reveals the framework's strength in creating evocative descriptions that prioritize meaningful atmosphere over a redundant list of objects.

\paragraph{Instance 3: Factual Accuracy and Contextual Reasoning}
An analysis of a street interaction(Figure \ref{fig:anli33333}) most clearly showcases TDSR's ability to correct errors and eliminate redundancy. The baseline description can include a long list of non-essential background details, representing significant informational redundancy. More critically, the baseline can falter on factual accuracy, for instance, by misidentifying a key object in the interaction. The TDSR model rectifies such errors, demonstrating superior object recognition. It also displays advanced contextual reasoning by interpreting body language and de-emphasizing the background to highlight the core interaction, resulting in a more focused and accurate narrative.

Collectively, these examples demonstrate that the TDSR framework systematically elevates image captioning from simple enumeration to rich, context-aware narrative generation. It consistently produces captions that are not only more detailed and insightful, but also more focused and factually reliable than those from the unenhanced baseline model.

\begin{figure*}[t] 

\begin{tcolorbox}[
    title=Prompt: Student Agent for Iterative Problem Refinement,
    colback=gray!5!white,
    colframe=gray!75!black,
    coltitle=white,
    fonttitle=\bfseries,
    arc=2mm,
    boxrule=0.8pt
]
\small
You are the Student Agent in the COGENT framework.  
Your role is to perform \textbf{Iterative Problem Refinement (IPR)} on a flawed mathematical problem, systematically externalizing your reasoning process into a clear and analyzable form.

\vspace{0.5em}
\textbf{1. Task Overview:}  
In \textbf{each iteration}, you will complete two explicit steps to process the given flawed problem.

\vspace{0.5em}
\textbf{2. Steps:}  
\begin{enumerate}[leftmargin=1.5em]
    \item \textbf{Step 1 – Critique:}  
    \begin{itemize}[leftmargin=1.2em]
        \item Carefully read the current problem statement in its entirety.
        \item Identify \textbf{all} logical flaws, missing conditions, contradictory statements, invalid assumptions, or ambiguous/misleading phrasings.
        \item For each flaw, explain \textit{why} it is problematic and \textit{how} it affects the solvability, clarity, or correctness of the problem.
        \item Consider both explicit errors (e.g., wrong numbers, impossible conditions) and implicit issues (e.g., missing definitions, unclear scope).
        \item Be precise and concise. \textbf{Do NOT} attempt to solve the problem; focus only on diagnosing and describing issues.
        \item Structure your critique as a clear list of independent, actionable points.
    \end{itemize}

    \item \textbf{Step 2 – Refine:}  
    \begin{itemize}[leftmargin=1.2em]
        \item Based strictly on your critique, rewrite the problem statement so that it becomes:
        \begin{enumerate}
            \item Logically consistent — all conditions align without contradictions.
            \item Complete — all necessary information and constraints are explicitly included.
            \item Well-posed — the problem can be solved unambiguously by a competent solver.
        \end{enumerate}
        \item Preserve the original intent of the problem as much as possible while fixing flaws.
        \item Avoid introducing new ambiguities or altering the intended difficulty level unless necessary for clarity.
        \item Keep language clear, formal, and precise, ensuring no room for misinterpretation.
    \end{itemize}
\end{enumerate}

\vspace{0.5em}
\textbf{3. General Instructions:}  
\begin{itemize}[leftmargin=1.5em]
    \item Always critique first, then refine — the refinement must be traceable to the identified issues.
    \item Keep the critique factual and objective; avoid speculative assumptions unless explicitly required to repair the problem.
    \item \textbf{Do NOT} provide the solution to the problem.
    \item Ensure the refined version is \textbf{self-contained}, understandable, and solvable without external context.
\end{itemize}

\vspace{0.5em}
\textbf{4. Required Output Format:}  
\begin{verbatim}
Critique:
<Your critique in bullet points or numbered list, one flaw per point, with explanations>

Refined Problem:
<Your revised problem statement ensuring clarity, completeness, and solvability>


\end{verbatim}

\end{tcolorbox}

\end{figure*}

\begin{figure*}[t]
\begin{tcolorbox}[
    title=Prompt: Teacher Agent for Iterative Problem Refinement,
    colback=gray!5!white,
    colframe=gray!75!black,
    coltitle=white,
    fonttitle=\bfseries,
    arc=2mm,
    boxrule=0.8pt
]
\small
\textbf{Role:} You are the \textbf{Teacher Agent} in the COGENT framework. Your role is to critically evaluate the Student Agent's critique and refined problem, and to produce an authoritative, pedagogically sound final version of the problem.

\vspace{0.5em}
\textbf{Instructions:}
\begin{enumerate}[leftmargin=1.5em]
    \item Read the flawed problem, the student's critique, and the student's refined problem statement.
    \item Verify whether the critique has correctly identified all major flaws; if important flaws are missing, add them.
    \item If the critique contains inaccuracies or invalid points, correct them.
    \item Based on the corrected critique, rewrite the problem statement such that it is:
    \begin{enumerate}
        \item Fully clear and unambiguous
        \item Free from any logical inconsistencies
        \item Complete with all necessary conditions for solvability
        \item Mathematically correct and well-posed
    \end{enumerate}
    \item While improving precision and clarity, preserve the original educational intent of the problem.
\end{enumerate}

\vspace{0.5em}
\textbf{Required Output Format:}
\begin{verbatim}
Teacher's Notes:
- Additional or corrected critique points
- Justification for any changes

Teacher's Refined Problem:
<Final, high-quality revised problem statement>
\end{verbatim}
\end{tcolorbox}
\end{figure*}

\end{document}